\documentclass[acmtog,screen,nonacm]{acmart}

\setcopyright{none}
\renewcommand\footnotetextcopyrightpermission[1]{}
\acmDOI{}
\acmISBN{}
\acmConference{}{}{}
\acmBooktitle{}
\acmYear{}
\copyrightyear{}

\usepackage[dvipsnames]{xcolor} 
\usepackage{graphicx}
\usepackage{tikzducks}
\usepackage{subcaption}
\usepackage{algorithm}
\usepackage{algpseudocode}
\usepackage{amsmath}
\usepackage{microtype}
\usepackage{cleveref}
\usepackage{soul}
\usepackage{xspace}
\usepackage{tabularx}
\usepackage{orcidlink}
\usepackage{comment}
\usepackage{xspace}
\usepackage{mathtools}
\usepackage{tabularx}
\usepackage{booktabs}
\usepackage{listings}
\usepackage{wrapfig}

\DeclareMathOperator*{\argmin}{arg\,min}

\newcommand{\name}{ViPS\xspace}

\newcommand{\changed}[1]{{#1}}
\newcommand{\honglin}[1]{{#1}}

\begin{document}

\title{\name: Video-informed Pose Spaces for Auto-Rigged Meshes}

\author{Honglin Chen}
\authornote{This project was done during Honglin's internship at Adobe Research.}
\orcid{0009-0005-9311-2355}
\affiliation{%
  \institution{Columbia University}
  \country{USA}
}

\author{Karran Pandey}
\orcid{0000-0002-4144-1705}
\affiliation{%
  \institution{University of Toronto}
  \country{Canada}
}

\author{Rundi Wu}
\orcid{0000-0003-0133-8196}
\affiliation{%
  \institution{Columbia University}
  \country{USA}
}

\author{Matheus Gadelha}
\orcid{0000-0002-4971-7980}
\affiliation{%
  \institution{Adobe Research}
  \country{USA}
}

\author{Yannick Hold-Geoffroy}
\orcid{0000-0002-1060-6941}
\affiliation{%
  \institution{Adobe Research}
  \country{Canada}
}

\author{Ayush Tewari}
\orcid{0000-0002-3805-4421}
\affiliation{%
  \institution{University of Cambridge}
  \country{United Kingdom}
}

\author{Niloy J. Mitra}
\orcid{0000-0002-2597-0914}
\affiliation{%
  \institution{Adobe Research}
  \country{United Kingdom}
}
\affiliation{%
  \institution{University College London}
  \country{United Kingdom}
}

\author{Changxi Zheng}
\orcid{0000-0001-9228-1038}
\affiliation{%
  \institution{Columbia University}
  \country{USA}
}

\author{Paul Guerrero}
\orcid{0000-0002-7568-2849}
\affiliation{%
  \institution{Adobe Research}
  \country{United Kingdom}
}
\renewcommand{\shortauthors}{Honglin Chen et al.}

\begin{teaserfigure}
 \includegraphics[width=\textwidth]{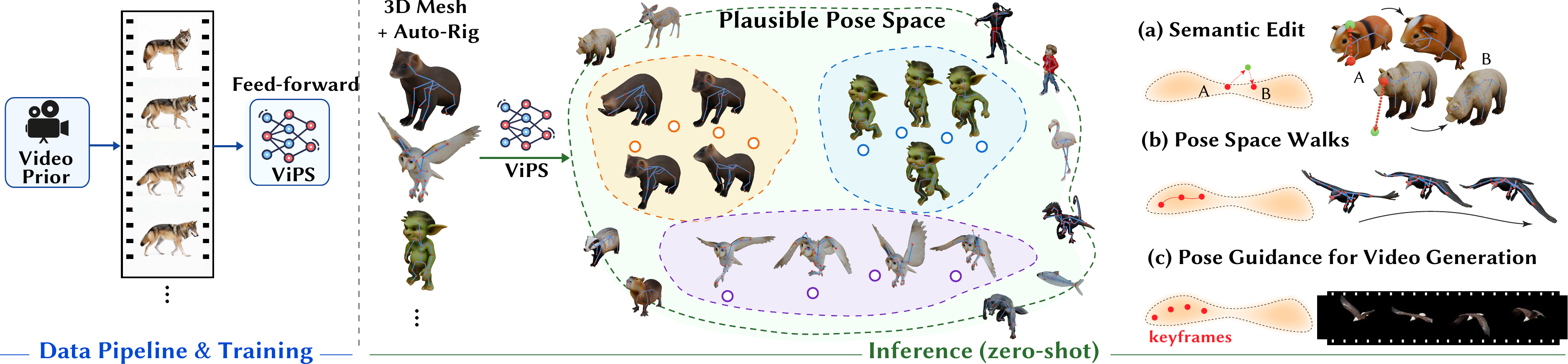}
  \caption{
  We introduce \name, a universal feedforward model that lifts static, auto-rigged meshes into a \textit{pose space}, i.e., a plausible and navigable manifold of deformed shapes. Specifically, we distill latent motion priors of foundational video models into a feedforward model to reveal such a pose space. This enables several applications: 
(a)~manifold-constrained editing; (b)~smooth pose-space interpolation, and (c)~pose-guided video synthesis. The discovered pose space remains query-able using a single 3D mesh and its autorig (using RigAnything~\cite{liu2025riganything}) to generate a manifold of plausible poses; invalid configurations, such as unnatural bone twisting, fall outside this manifold (indicated with red dots). Please see supplemental videos and 4D visualizations. }
  \label{fig:teaser}
\end{teaserfigure}

\begin{abstract}
Kinematic rigs provide a structured interface for articulating 3D meshes but lack any associated \textit{pose space}, i.e., an explicit representation of the \textit{plausible} manifold of joint configurations for a given mesh. Without such a pose space, stochastic sampling or manual manipulation of raw rig parameters easily results in semantic and/or geometric violations, such as anatomical hyperextension and non-physical self-intersections. We propose Video-informed Pose Spaces (\name), a feedforward framework that discovers the latent distribution of valid articulations for (auto-)rigged meshes by distilling motion priors from a pretrained video diffusion model. 
Unlike existing methods that rely on scarce, artist-authored 4D datasets, or focus on reconstructing instances of individual motions, \name transfers generative video model priors into a universal distribution over the given rig parameterization. Differentiable geometric validators applied to the skinned mesh enforce shape-specific integrity without requiring manual regularizers. Our feedforward model reveals a smooth, compact, and controllable pose space. This, in turn,  supports sampling for diverse shape variations, manifold projection for inverse kinematics, and temporally coherent trajectories for animation and keyframing. 
Further, the distilled 3D pose samples serve as semantic proxies to guide video diffusion, effectively closing the loop between generative 2D priors and structured 3D kinematic control. Our evaluations show that \name, trained solely using video priors, matches the performance of state-of-the-art models trained on synthetic artist-created 4D data in both plausibility and diversity. Additionally, as a universal model, \name exhibits robust zero-shot generalization to out-of-distribution species and unseen skeletal topologies. Project Page: \href{https://honglin-c.github.io/vips/}{\texttt{https://honglin-c.github.io/vips/}}

\vspace{-5pt}

\end{abstract}

\maketitle

\section{Introduction}

Kinematic rigs, such as shape skeletons, turn a 3D shape into an editable asset by exposing a set of controls joints. While this yields a low-dimensional parameterization of motion, it does not immediately enable finding \textit{plausible poses} in that space. 
Naively exploring the parameters of such a (skeleton) rig readily produces implausible states such as hyperextension, unnatural twisting, or semantically implausible poses.
In practice, artists manually author pose spaces by restricting and correlating rig parameters, ensuring edits result in plausible, artifact-free deformations.
Although a powerful tool in the hands of skilled artists, a rig alone, without associated bounds and/or parameter coupling, does not specify \emph{where} in this control space an asset can plausibly go.
Modeling a \emph{plausible pose space} (in short, \textit{pose space}) for a given 3D shape (i.e., the distribution of joint configurations that are valid and semantically consistent for the shape) and the ability to traverse it make pose edits aware of asset semantics. Edits can be projected into the pose space and one can create asset-specific animations as walks through the pose space (see Figure~\ref{fig:teaser}).
Unfortunately, such a pose space
is not straightforward to define. 
In this work, we ask: \emph{Given an auto-rigged 3D mesh, can we automatically discover its (plausible) pose space?}

A key obstacle to answering this question affirmatively is the lack of supervision data.
Learning mesh-specific pose distributions ideally requires a large repository of 3D/4D articulated data in the form of rigged meshes animations or pose sets, spanning the range of plausible configurations per asset.
Such data is scarce, expensive to curate at scale, and typically category-specific, limiting its applicability across the long tail distribution of shapes.
Meanwhile, modern video diffusion models~\cite{wiedemer2025veo,AdobeFirefly2026,Sora2,yang2024cogvideox, wan2025} are purportedly trained on vast video corpora to encode strong, scalable priors over plausible motion and pose.
Yet these priors are not grounded in a particular skeleton/rig and offer no precise, low-dimensional control interface.
Thus, existing options provide either \emph{controls without plausibility} (rigging) or \emph{plausibility without control} (video priors).
Furthermore, while 4D reconstruction methods~\cite{puppeteer, s3o, magicpose4d} can recover rigged 4D motion sequences from videos,
they remain fundamentally `instance-based.' They do not learn the underlying manifold of plausible poses, leaving them unable to support interactive applications like semantically-aware pose editing through pose space projection, or smooth pose space traversal. Consequently, there remains a need for a representation that not only captures motion but organizes it into a navigable, generative domain. 

\begin{figure}[t]
    \vspace{-5pt}
    \centering
    \includegraphics[width=\linewidth]{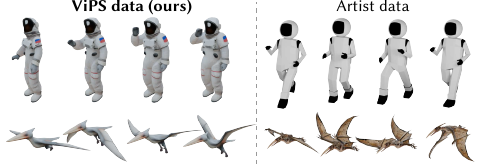}
    \caption{
    \textbf{Data comparison between artist-authored data and our data.} 
    Learning controllable pose spaces requires large amounts of articulated 3D/4D motion data.
    Free, open-source datasets such as Objaverse-XL~\cite{objaverseXL} (top right) often contain animations whose quality varies substantially depending on asset origin and animator effort, while commercial datasets such as Truebones Zoo~\cite{trueboneszoo} (bottom right) can provide higher-quality animations but are typically smaller in scale and expensive to acquire.
    In contrast, our data generation pipeline (left) converts rich motion priors from video models into diverse, correspondence-aware 4D motions, providing scalable supervision for learning navigable pose spaces.
    } 
    \label{fig:supp_artist_data_comparison}
    \vspace{-5pt}
\end{figure}

We propose Video-informed Pose Spaces~(\name), which \emph{distills video diffusion priors into a mesh-conditioned pose space} that can be sampled using a universal feed-forward network, which, in turn, unlocks different ways to semantically and precisely drive mesh deformation as well as video generation.
Given a mesh and its rig~\cite{xu2020rignet,liu2025riganything} as condition, \name models a generative distribution over rig parameters (i.e., our pose space) using a diffusion model in rig space.
We do \textit{not} require any articulated 3D/4D repositories as supervision to train \name. 
Instead, we directly leverage a video diffusion model (TurboDiffusion Wan 2.2~\cite{TurboDiffusion2025} in our tests)
as a prior over plausible pose evolution, which we distill into the rig-parameter distribution.
To robustly transform a video signal into rig-parameter space, we optimize rig parameters to match 4D motion guidance of the video based on ActionMesh~\cite{sabathier2026actionmeshanimated3dmesh}, coupled 
with geometric priors applied directly to the posed, skinned mesh, encouraging samples that are both semantically plausible and geometrically valid for a given shape (see Figure~\ref{fig:supp_artist_data_comparison}). 
This video-informed data is then used to train our generative pose space model.

The resulting pose space is not only generative but also readily \emph{navigable} and, empirically smooth. 
This low-dimensional distribution in rig space enables (i)~\emph{sampling} a variety of valid, plausible poses; (ii)~\emph{projection} of arbitrary or user-edited joint configurations back onto the manifold of plausible poses for handle-driven deformation (and inverse kinematics~(IK)) as well as DDIM inversion with image diffusion; 
and (iii)~\emph{traversal} that traces coherent pose trajectories for animation and keyframing. 
These generated poses and trajectories can be directly used to animate meshes or rendered as \emph{animated shape proxies} that provide stable, semantically aligned 3D guidance for image or video diffusion models. 
See \Cref{fig:teaser}.

We evaluate \name across an extensive suite of diverse deformable assets, spanning a wide array of biological species and skeletal topologies. Our experiments demonstrate that by distilling video-informed priors, our model matches the performance of state-of-the-art architectures~\cite{gat2025anytop} trained on synthetic 4D data (e.g., Objaverse-XL, TrueBones) in terms of both pose plausibility and distributional diversity. Further, as a universal generative model, \name exhibits zero-shot generalization to unseen species, successfully discovering valid pose spaces for assets entirely absent from the training distribution. We quantitatively validate these findings through metrics for pose validity, manifold coverage, and temporal smoothness, while showcasing superior controllability in downstream tasks. We observe similar behavior when distilling other video models (see supplemental material).

In summary, our main contributions are:
(i)~formulating \emph{pose space discovery} as learning a universal mesh-conditioned generative distribution over rig parameters;  unlike 4D reconstruction recovering specific instances of motion, \name learns the continuous manifold of valid configurations and enables applications like semantic pose edits and pose space walks;
(ii)~distilling video-to-pose space through a video diffusion model to supervise and transfer motion priors into rig space \textit{without} curated 3D/4D motion/pose data to handle the long tail of shape variations; and 
(iii)~introducing a high-quality 4D motion dataset with correspondence, containing 127k poses spanning 100+ species and 200+ unique individuals built from generative video priors with VLM guidance and 4D reconstruction.

\section{Related Work}
\label{sec:related}

\begin{figure*}[t]
    \centering
    \includegraphics[width=\linewidth]{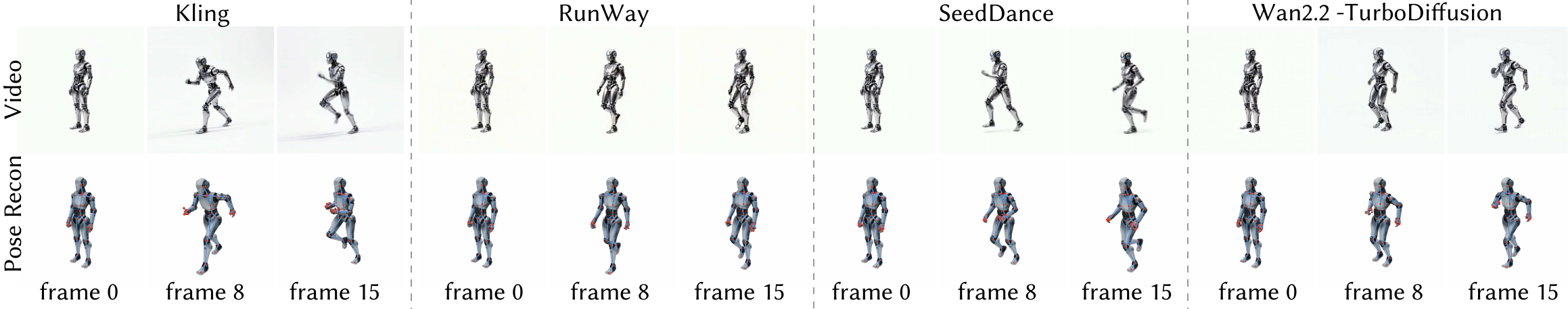}
    \caption{\textbf{Different video model priors.} \honglin{
    Our data pipeline can be seamlessly integrated with different video models as priors. Here, we show pose optimization results obtained by switching from our current video model Wan2.2-TurboDiffusion~\cite{TurboDiffusion2025} (right) to other video models (Kling~\cite{klingteam2025klingomnitechnicalreport}, RunWay~\cite{runway2026} and SeedDance~\cite{seedance2026seedance20advancingvideo}), while using the same input image and text prompt describing a robot running.
    }
    }
    \label{fig:supp_different_video_priors}
\end{figure*}

\paragraph{Encoding animations.}
A central question in character animation is \emph{how to parameterize motion}. One naive approach is to encode deformation directly as \emph{mesh sequences} \cite{lengyel1999compression}, which is expressive but high-dimensional and costly to store, edit, and generalize.
Instead, artists rely on low-dimensional \emph{rig-based} parameterizations, including linear blend skinning (LBS) \cite{lewis2000lbs} and dual quaternion skinning \cite{kavan2008dual}, often augmented with energy-based regularizers that preserve local rigidity or volume under articulation \cite{kmp_shape_space_sig_07,lipman_volPreserving_07}.
Alternative formulations such as cage-based deformation and differential coordinates \cite{yu2004mesh,sumner2004deformation} similarly provide structured controls through geometric constraints.
More recently, PhysRig \cite{zhang2025physrig} proposes a differentiable physics-based alternative to LBS, modeling soft-body volumes to capture complex dynamics like tissue jiggling. In the context of humans, Pose-NDF \cite{tiwari22posendf} models the manifold of plausible human poses as a zero-level set of a Neural Distance Field (NDF) in a joint configuration space. While effective, such implicit manifolds are often category-specific (i.e., humans) and lack the direct interpretability of a rig-based latent distribution. 
While these encodings, when carefully authored, can produce visually convincing motion, they typically require manual rigging. Further, they do not by themselves specify \emph{what values are valid}, i.e., they expose degrees of freedom but not the bounded, coupled, semantically plausible \emph{pose space}.  
The resultant movements remain semantic-agnostic, failing to capture the manifold of natural poses for any arbitrary shape inherent to its specific biological or mechanical structure.

\paragraph{Learning rigs from artist data.}
Learning-based auto-rigging turns static meshes into controllable assets by inferring rigs and deformation models from curated, artist-authored data.
RigNet \cite{xu2020rignet} pioneered supervised rig prediction using collections of rigged models with skeletons and skin weights, and subsequent systems infer skeletons, part structures, and deformation weights more broadly, including in weaker supervision regimes~\cite{articulateanymesh,deng2025anymate,digitaltwinart}; see also articulated-object (e.g., furniture) variants \cite{li2025particulate,jiayi2024singapo}.
These methods effectively recover an \emph{efficient parameterization} (rig + weights) and thus a low-dimensional \emph{control interface}, but they typically stop short of learning the \emph{distribution} over controls: joint limits, parameter couplings, and the likelihood of poses implied by real motion. Concurrent to ours, RigMo \cite{zhang2026rigmo} jointly learns rig structure and motion dynamics directly from raw mesh sequences via a dual-path VAE. However, RigMo remains dependent on available 4D mesh sequences for training. Instead, we shift the supervision from 3D/4D artist data to the scalable prior of video diffusion models, enabling the discovery of pose spaces for assets where no ground-truth motion sequences exist.

\paragraph{Reconstructing 4D from video data.}
A separate line of work, inspired by recent breakthroughs in computer vision and tracking, obtains 4D (space--time) content from video by reconstructing \emph{renderable} dynamic representations, most commonly via per-sequence optimization or amortized prediction.
\textit{Optimization-based methods} fit time-varying meshes or auto-rigs (e.g., Puppeteer~\cite{puppeteer}, {S3O~\cite{s3o} and MagicPose4D~\cite{magicpose4d}}), dynamic NeRFs, or dynamic point-based renderers (e.g., Gaussian splats), producing high-fidelity reconstructions but requiring expensive test-time optimization, careful temporal regularization, and often canonical-template assumptions to stabilize correspondences \cite{Pumarola_2021_CVPR,Park2021HyperNeRF}.
Likewise,  4D Gaussian splatting systems improve rendering efficiency, yet still commonly rely on multi-stage per-video optimization and regularization to maintain stable motion, visibility, and appearance \cite{luo2025instant4d,liu2025mono4dgshdr,4dfly_wu_25}.
More recently, generative 
\textit{feed-forward} approaches amortize inference into a single forward pass \cite{Wei2024M2V,Jiang2025Geo4D,xu20254dgt,Ren2024L4GM,ShapeGen4D,gvfdiffusion,ss4d}, but their outputs are typically implicit fields, point-/splat-based primitives (e.g., 4D Gaussians) rather than a semantic, low-dimensional control space.
We utilize 
ActionMesh~\cite{sabathier2026actionmeshanimated3dmesh}, a feed-forward model that produces a displacement field to enable converting videos to animated meshes with tracked vertices. 
Utilizing this information, we develop the missing layer: instead of reconstructing 4D appearance, we distill video priors into plausible \emph{pose space} to learn a structured latent distribution to sample from, enabling meaningful pose space navigation. 

\begin{figure*}[t]
    \centering
    \includegraphics[width=\linewidth]{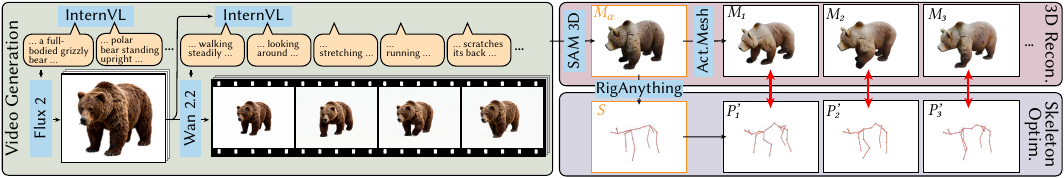}
    \caption{\textbf{The \name data pipeline}. Given a species like `\texttt{bear}', we first generate a clean single-object image of diverse species representatives using an image generator with prompts generated by a carefully instructed VLM. We then expand these images into many videos of diverse motions using a VLM-prompted video generator. We use a VLM to choose a frame $\alpha$ as rest pose, reconstruct a textured mesh from the frame using SAM3D~\cite{chen2025sam3d} and obtain meshes with shared topology for all other frames with the recent ActionMesh~\cite{sabathier2026actionmeshanimated3dmesh}. We generate a skeleton for frame $\alpha$ using RigAnything~\cite{liu2025riganything} and find skeleton poses for the other frames by optimizing the skeleton to match their 3D reconstructions. Using a video prior enables a more scalable pipeline to generate diverse motions and objects without relying on artist-created 3D/4D animation datasets.}
    \label{fig:data_pipeline}
\end{figure*}

\paragraph{Generative motion models beyond reconstruction.}
Skeletal-free methods generate motion directly as point clouds~\cite{mo_24_motionKeyframe}, shape-handles~\cite{zhang2024tapmo}, or deforming meshes~\cite{muralikrishnan2025smftemplatefreerigfreeanimation,ye2024skinned,zhang2022motiondiffuse}.
These representations offer flexibility across character topologies, but they forgo the key benefits of skeletons: rigs provide a compact, semantically meaningful interface that is easy to manipulate with standard rig-based animation tools, and integrates naturally with animation workflows and inverse-kinematics pipelines. A separate class of models focuses on the generative capabilities of pose spaces. DPoser-X \cite{lu2025dposerx} leverages diffusion to model whole-body human priors, enabling robust pose completion from partial observations (e.g., inferring upper-body poses from legs); for quadrupeds, QuadForecaster \cite{noronha2025quadforecaster} utilizes cascaded diffusion for movement forecasting in animal communication analysis. While these models excel at individual categories, Dimo \cite{mou2025dimo} generalizes this by distilling video priors into 3D motion for arbitrary objects. %
At the other end, methods that explicitly support arbitrary skeletons often require \emph{topology-specific training} (either separate model instances or skeleton-specific adaptations), which scales poorly and limits cross-skeleton generalization~\cite{li2022ganimator,zhang2023skinned}. 
AnyTop~\cite{gat2025anytop} is the closest skeletal-based baseline: it introduces a single diffusion model conditioned on skeletal topology and joint names, designed to generalize across diverse topologies; making it a strong baseline for multi-class motion generation, though it relies on curated 4D training data and does not explicitly discover a rig-parameter pose space from raw rigs. 
Additionally, both text labels and motion statistics for each joint need to be given as input (the mean and variance over joint motions are required for normalization), limiting its applicability to unseen species/shapes.

\section{Method}
\label{sec:method}
Given a textured mesh $M$ (with $N_V$ vertices) rigged with skeleton 
$S = (P, E, W)$, where $P\in \mathbb{R}^{N_P \times 3}$ are skeleton node positions, $E \in \mathbb{N}^{N_E \times 2}$ are skeleton edges, and $W \in \mathbb{R}^{N_P \times N_V}$ are skinning weights,
our goal is to model the conditional distribution of plausible skeleton node positions $p(P' | M, S)$, corresponding to the distribution of plausible poses $P'$, with a feed-forward network $f_\theta$. Datasets with artist-rigged and artist-animated meshes are scarce, while video priors can create much more plentiful and varied examples of plausible object poses; so we instead use a video prior to obtain training data for $f_\theta$. We first describe how to obtain rigged meshes with multiple plausible poses from a video prior, and then how we use this data to model the conditional distribution of plausible rig parameters with the feed-forward network $f_\theta$ and sample from it.

\subsection{Rigged Mesh Poses from a Video Prior} 
Our dataset consists of tuples $(M, S, P')$, where $P'$ describes a plausible \emph{deformed} pose of the rigged mesh $(M, S)$ that is different from the \emph{rest} pose $P \in S$. To obtain this dataset, we carefully prompt the video prior to generate videos of single objects performing various motions. Each frame of these videos corresponds to a plausible pose of the object. We then reconstruct an animated sequence of 3D meshes for each frame that are roughly aligned in a common world space. We use the first frame in each video as mesh $M$ and extract a skeleton $S$ from its 3D reconstruction using an existing method~\cite{liu2025riganything}. Finally, we optimize the pose parameters of $S$ to match each of the other frames, giving us plausible pose parameters $P'$ and thus a dataset tuple $(M, S, P')$ for each frame.
The data generation pipeline is shown in Figure~\ref{fig:data_pipeline}, each step is described in detail below.

\paragraph{Video Generation.}
To create videos of single objects performing various motions, the main difficulties are to (i)~obtain videos of single objects that have favorable conditions for 3D reconstruction, such as avoiding additional background or scene elements, good lighting, having the full object in frame, etc.; and (ii)~obtain videos of these objects performing a diverse set of plausible motions. To address the first issue, we split video generation into two steps: text-to-image generation with Flux 2~\cite{flux-2-2025}, followed by expanding this first frame into a video using Wan2.2~\cite{wan2025} (accelerated with TurboDiffusion~\cite{TurboDiffusion2025}). Note that the rest of our data pipeline is not tied to a specific video model and can leverage different video models as motion priors (see Figure~\ref{fig:supp_different_video_priors}). We found that creating the first frame with an image prior yields higher-quality videos and better prompt adherence for the first frame. We take advantage of this prompt adherence by automatically instructing the VLM InternVL~\cite{wang2025internvl3_5} to generate prompts that result in favorable conditions for 3D reconstruction. To address the second issue, we give InternVL the first frame and instruct it to generate detailed structured prompts for the video generator that describe diverse motion for the animal/object shown in the first frame. See the supplement for the exact instructions we give to InternVL in these two steps. We generate multiple videos for a given first-frame image, each with different motion descriptions generated by InternVL.

\paragraph{4D Reconstruction.}
Given a video, we reconstruct a textured 3D mesh that is then deformed/tracked over the frames. This is a 4D reconstruction task and we use Actionmesh~\cite{sabathier2026actionmeshanimated3dmesh} to obtain a sequence of meshes with shared mesh topology. However, as these meshes are untextured by default, we initialize Actionmesh with a SAM3D~\cite{chen2025sam3d} reconstruction of a single frame. This gives us higher-quality textured meshes $(M_1, M_2, \dots)$ with shared mesh topology across frames.
The choice of frame $\alpha$ for the initial mesh $M_\alpha$ is important, as different frames may produce different mesh topologies. For example, using a frame that shows a bird with closed wings results in a mesh topology that does not support opening the wings, as the underside of the wings is not represented in the mesh. Similarly, choosing a frame with large occluded parts can result in poor mesh quality. We instruct InternVL to choose a frame where the object is clearly visible and has minimal self-occlusions or motion blur.

\paragraph{Skeleton Optimization.}
To construct our dataset tuples $(M, S, P')$, we start by defining the initial mesh as a rest pose and set $M \coloneq M_\alpha$. We obtain the skeleton $S$ by running the RigAnything~\cite{liu2025riganything} skeleton extractor on $M$. For deformed poses $P'$, we need to share the same mesh and skeleton topology as the rest pose $(M, S)$, and thus cannot simply extract skeletons for each mesh $M_i$. Instead, we obtain $P'$ by optimizing the skeleton parameters of $S$, so that the skeleton-deformed mesh $M'$ matches the reconstructed mesh $M_i$ as closely as possible, by minimizing the following energy:
\begin{equation}
    \mathcal{L}_\text{recon}\ \coloneq\ d_\text{vert}(M_i, M')\ \text{with}\ M' \coloneq h(M, S, P'),
\end{equation}
where $h(M, S, P')$ denotes a deformation of mesh $M$ through the skeleton $S$ with node positions $P'$ using linear blend skinning~\cite{magnenat1989lbs, lewis2000lbs} and $d_\text{vert}$ is the mean square distance between corresponding vertices of two meshes. Recall that all meshes $M_i$ share topology, therefore corresponding vertices are known.
We include a skeleton regularization to encourage edge lengths in the skeleton to remain constant:
\begin{equation}
    \mathcal{L}_\text{edge}\ \coloneq\ \frac{1}{N_E} \sum_{N_E}  \left| \|P_{E_{i,0}} - P_{E_{i,1}}\|\ - \|P'_{E_{i,0}} - P'_{E_{i,1}}\| \right|,
\end{equation}
where $(E_{i,0}, E_{i,1})$ is the pair of node indices for skeleton edge $i$. The full optimization of the deformed pose $P'$ is then defined as:
\begin{equation}
    \argmin_{P'}\ \mathcal{L}_\text{recon} + \lambda \mathcal{L}_\text{edge},
\end{equation}
where we set the regularization weight $\lambda = 20$ in our experiments.

\subsection{Learning a Pose Space}
\label{sec:learning_pose_space}
We train a diffusion model~\cite{ho2020ddpm} to represent the conditional distribution $p(P' | M, S)$ using the dataset tuples $(M, S, P')$. The feed-forward network $f_\theta$ acts as the denoiser, taking as input a noisy version $P^t$ of the node positions $P'$, with noise level $t \in [0, T]$, and predicting the denoised node positions.
It is trained with a standard MSE loss:
\begin{equation}
    \mathcal{L}_\text{diff} \coloneq w(t) \left\|\ f_\theta(P^t; E, P, F, t) - P'\ \right\|_2^2,
\end{equation}
where $w(t)$ is a weighting schedule for time steps $t$. To make the denoiser aware of the skeleton topology and the rest pose, we additionally provide the skeleton edges $E \in S$ and the rest pose $P \in S$ as input. Further, for the mesh $M$, we provide semantics in the form of per-node semantic features $F$, as defined later.

Once trained, we sample poses $P^0 \sim p(P' | M, S)$ with the denoiser $f_\theta$ using a stochastic denoising strategy~\cite{ho2020ddpm}:
\begin{align}
    P^T &\sim \mathcal N(\mathbf 0, \mathbf I) \nonumber \\
    P^{t-1} &\sim \mathcal{N}\left(a_t P^t + b_t (f_\theta(P^t; E, P, F, t) - P^T), \sigma^2_t\mathbf{I}\right),
    \label{eq:inference_step}
\end{align}
where $a_t$, $b_t$, and the variance $\sigma^2_t$ are chosen according to a denoising schedule, and $\mathcal{N}$ is the normal distribution.

\begin{figure}[t]
    \vspace{-5pt}
    \centering
    \includegraphics[width=\linewidth]{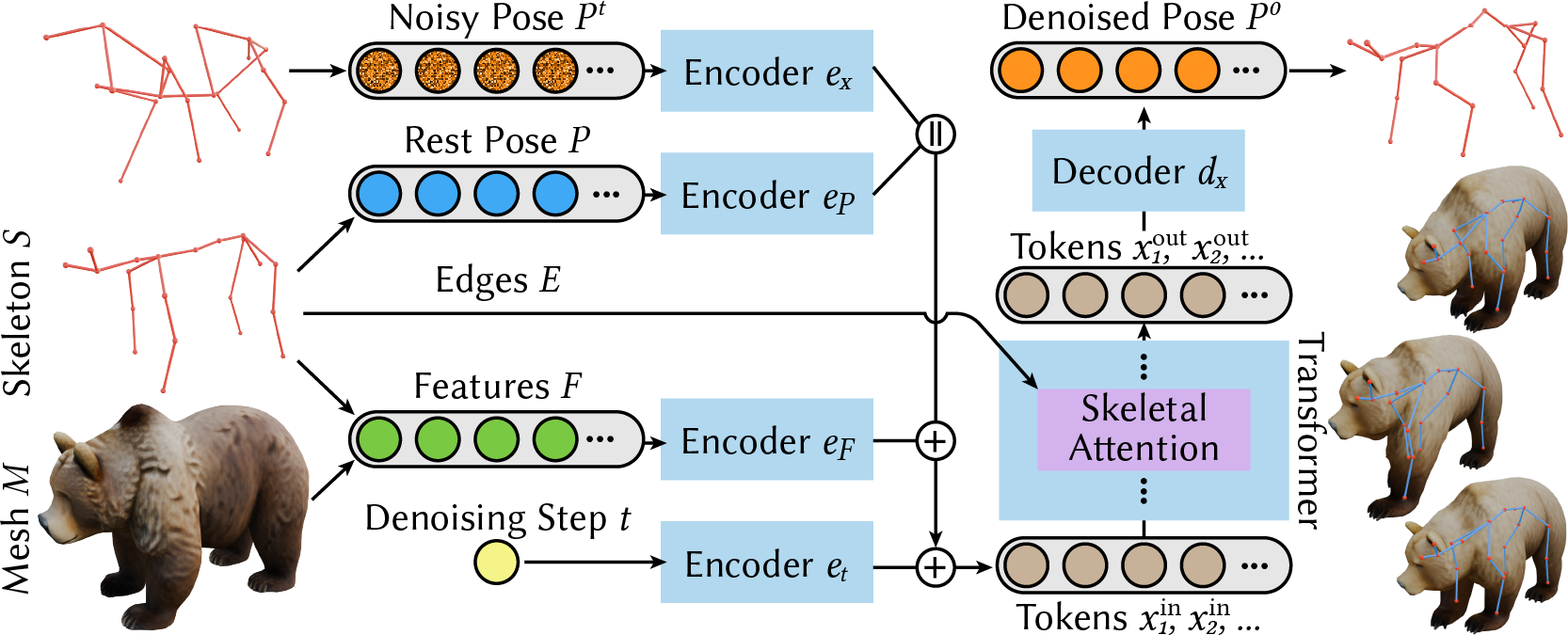}
    \caption{\textbf{The architecture of our universal pose  denoiser $f_\theta$}. Given a mesh $M$ and its skeleton $S$ in rest pose, we sample plausible poses using multiple iterations of the denoising process illustrated here. The rest pose is concatenated as reference to the noisy pose, and to capture skeleton node semantics, we compute per-node semantic features from the textured mesh. This information is aggregated into a set of per-node tokens and used to estimate a denoised pose with a transformer. Skeleton edges are used to guide attention using a specialized \emph{Skeletal Attention} block~\cite{gat2025anytop}.}
    \label{fig:architecture}
\end{figure}

\begin{figure*}[t]
    \vspace{-10pt}
    \centering
    \includegraphics[width=\textwidth]{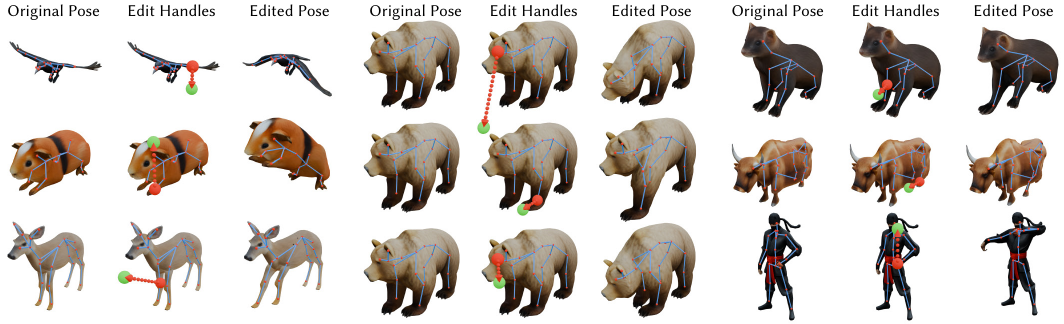}
    \caption{
    \textbf{Manifold-Constrained Semantic Editing.} 
    \name enables precise Inverse Kinematics (IK) by projecting user-driven joint handles (orange$\rightarrow$green) into the discovered \textit{plausible pose space}. By leveraging our \textit{video-informed priors}, our model ensures that sparse edits result in anatomically valid and structurally consistent poses across a wide range of species and skeletal topologies, effectively avoiding the geometric artifacts common in unconstrained rig manipulation.}
    \label{fig:pose_editing}
\end{figure*}

\paragraph{Semantic Node Features.}
In addition to the skeleton, our denoiser requires information about the mesh $M$ to accurately model the distribution of plausible poses. A dog and a cat, for example, may have similar skeletons, but different pose distributions. We provide information about the mesh as a set of per-node semantic features $F \in \mathbb{R}^{N_P\times N_F}$, with feature dimension $N_F$. Each feature $F_i \in \mathbb{R}^{1 \times N_F}$ describes the semantics of the part in mesh $M$ that is controlled by skeleton node $i$. Specifically, we first use Diff3f~\cite{dutt2024diff3f} to obtain semantic features $F^V \in \mathbb{R}^{N_V \times N_F}$ for each mesh vertex, where $N_V$ is the number of vertices in $M$. By default, Diff3f computes these features as a combination DINO-based features~\cite{oquab2024dinov2} and features from the intermediate layers of a 2D diffusion model. Both are obtained from multi-view renders of the mesh with generated texture. As our meshes $M$ are already textured, we use the DINO-based features of Diff3f from our texture and the diffusion features of the generated texture, where the former offers the semantics provided by DINO and the latter contains spatial information provided by the diffusion features. Finally, we use the skinning weights $W$ of the skeleton $S$ to aggregate the per-vertex features $F^V$ into per-node features $F$:
\begin{equation}
    F_i \coloneq \frac{\sum_{j=1 \dots N_V} W_{i,j} F^V_j}{\sum_{j=1 \dots N_V} W_{i,j}}. 
\end{equation}

\paragraph{Denoiser Architecture.}
Since the number of nodes in a skeleton is variable, we use a diffusion transformer (DiT) that is capable of processing a variable number of tokens as backbone of the denoiser $f_\theta$. An overview of the architecture is shown in Figure~\ref{fig:architecture}. We base our architecture on AnyTop~\cite{gat2025anytop} that we modify to (i)~model a distribution of poses rather than a distribution of pose sequences by removing the time dimension and the temporal attention; (ii)~use our semantic node features $F$ instead of a T5-encoded~\cite{raffel2020t5} text label for each node, allowing us to apply the method to unlabelled skeletons; and (iii) use a different normalization for node positions $P'$ that removes the requirement to provide a ground truth example motion for each input skeleton.
Specifically, we work with node positions $P'$ that are normalized as $(P' - P) / \sigma_P$, that is, we use the offset of nodes from their rest pose, normalized by the dataset-wide average offset of any node from its rest pose $\sigma_P$.
First, we encode all inputs $(P^t, P, F, t)$ except the skeleton edges $E$ into a token sequence $x^\text{in}_1, \dots, x^\text{in}_{N_P}$ that can be processed by the DiT, with one token per node. Each input is encoded by a separate linear layer, and aggregated into tokens:
\begin{equation}
    x^\text{in}_i \coloneq \left(e_x(P^t)\ |\ e_P(P)\right) + e_F(F) + e_t(t),
\end{equation}
where $|$ denotes concatenation and $e_*$ denote the encoders. These tokens are then processed by the DiT:
\begin{equation}
    x^\text{out}_1\ \dots x^\text{out}_{N_P} \coloneq \text{DiT}(x^\text{in}_1, \dots, x^\text{in}_{N_P};\ E),
\end{equation}
where the skeleton edges $E$ are used in a special \emph{Skeletal Attention} mechanism that makes the attention computation for pairs of nodes dependent on their topological relationships. See AnyTop~\cite{gat2025anytop} for more details. The output tokens $x_i^\text{out}$ are then decoded by a small feed-forward network $d_x$ :
\begin{equation}
    f_\theta(P^t; E, P, F, t) \coloneq d_x(x^\text{out}_1\ \dots x^\text{out}_{N_P}).
\end{equation}

\subsection{Constrained Sampling of the Pose Space}

\paragraph{Inverting/reconstructing a given pose.}
We approximately reproduce a given pose $P'$ by inverting it into a noise sample $P^T$ using DDIM inversion~\cite{song2021ddim}. Starting the inference process in \Cref{eq:inference_step} from this noise sample approximates the pose $P'$.

\paragraph{Sampling with sparse constraints.}
We can generate a pose that follows our learned prior while also approximately following simple constraints using a guided sampling strategy~\cite{pandey2025motion}, similar to classifier-free guidance~\cite{ho2021cfg}. Constraint violation is modeled with an energy function $\mathcal{E}$, and in each inference step, the denoising trajectory is nudged to minimize the energy term by changing the inference procedure in \Cref{eq:inference_step}:
\begin{align}
    P^{t-1} &\sim \mathcal{N}\left(a_t P^t + b_t (f_\theta(\widehat{P}^t; E, P, F, t) - P^T), \sigma^2_t\mathbf{I}\right) \text{, with} \\
    \widehat{P}^t &\coloneq P^t - \nabla_{P^t} \mathcal{E}\left(f_\theta(P^t; E, P, F, t) \right). \nonumber
\end{align}

\section{Results}
\label{sec:results}

\begin{figure*}[t]
    \centering
    \includegraphics[width=\linewidth]{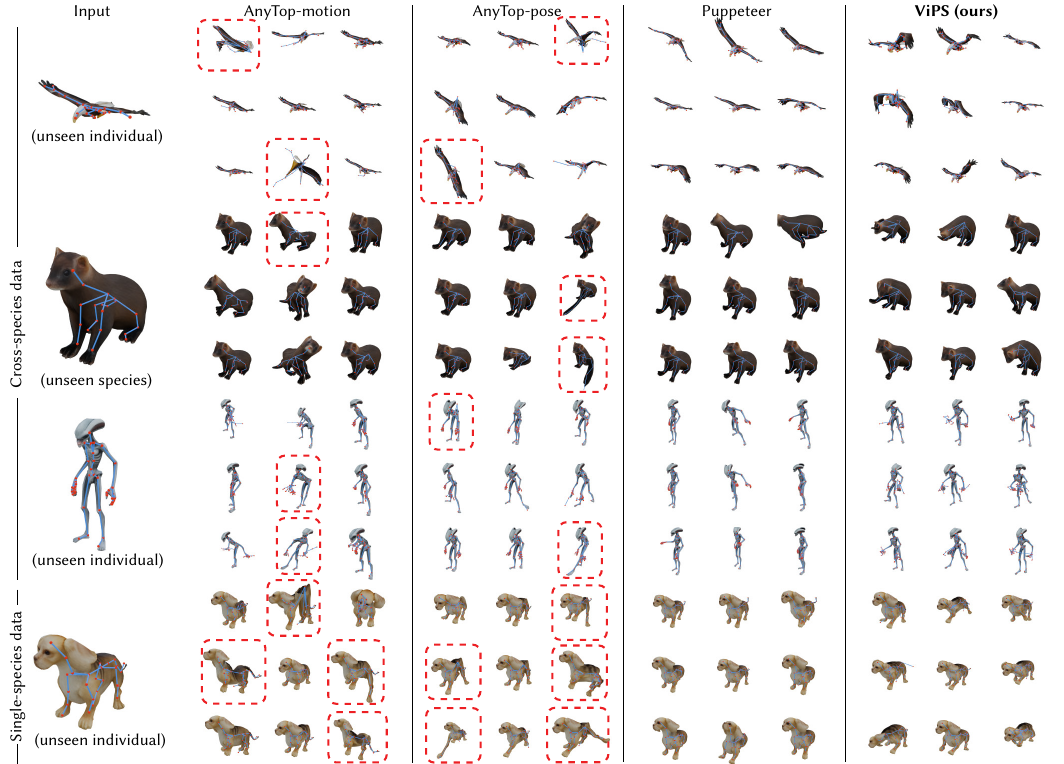}
    \caption{
    \textbf{Qualitative Comparison.} 
    We compare random poses samples generated with \name to all baselines on three individuals of the cross-species dataset and one individual of the single species dataset. Strongly implausible poses are marked in red. Our data generation allows generalization to a large range of species, such as the alien, and gives pose plausibility and diversity on par with the much more expensive Puppeteer.}
    \label{fig:qual_comparison}
    \vspace{5pt}
\end{figure*}

\begin{figure*}[t]
    \vspace{5mm}
    \centering
    \includegraphics[width=\linewidth]{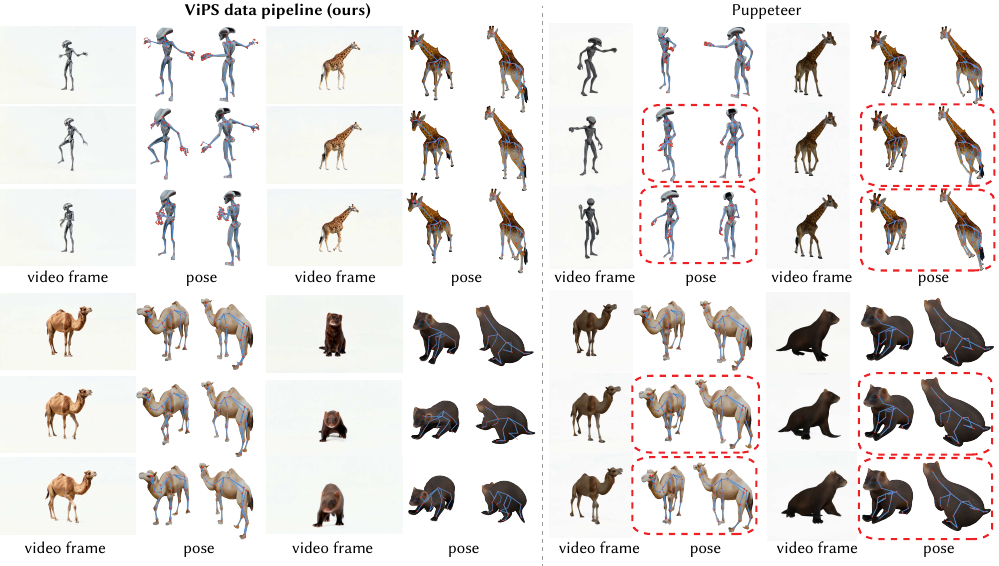}
    \caption{\textbf{Data comparison.} We compare our generated data to Puppeteer independently of the feed-forward model. Both our pipeline and Puppeteer reconstruct poses from video frames.
    In these examples, we fit poses to frames of a video that was generated from scratch using our data pipeline, while Puppeteer requires a video that was generated using a render of the rest pose mesh $M_\alpha$ as initial frame.
    For each generated pose (rendered from two different views), we also show the video frame that the pose was reconstructed from. We can see that Puppeteer often fails to accurately reconstruct limb poses (highlighted in red). We found that this is usually caused by tracking errors between frames, especially after self-occlusion or large motions. Our 4D reconstruction does not require explicit tracking or inter-frame correspondences, resulting in more pose reconstructions that more accurately follow the video frames.}
    \label{fig:supp_data_comparison}
\end{figure*}

We evaluate \name by comparing randomly sampled poses
to baselines, and demonstrate semantic pose editing as application.
Videos, a larger set of results, experiments with different video priors, and additional applications such as keyframe-controlled video generation and latent space interpolation are available in the supplemental document and webpage: \href{https://honglin-c.github.io/vips/}{\texttt{https://honglin-c.github.io/vips/}}.

\paragraph{Training and Data Generation.}
We train our model for 500k iterations on a single NVIDIA RTX A6000 GPU for three days. We use 100 diffusion steps, a batch size of 32, and a latent dimension of 128. For data generation, we run our data generation pipeline on 32 NVIDIA A100 GPUs for two days. Code will be made available.

\paragraph{Datasets.}
We evaluate on two types of human and animal data sets: A \texttt{cross-species} dataset focusing on diversity across species, and a \texttt{single-species} dataset focusing on diversity of individuals in a single species.
\textit{This data will be released on publication.} 

For the \texttt{cross-species} training set, we select 100 random species that include mammals, birds, fish, humans, and fantasy creatures, and generate 2 images for each, corresponding to two individuals from the species with distinct appearance (for example a \texttt{Puppy} and a \texttt{Labrador} dog).
Using 20 different motion prompts per individual, we generate 20 diverse videos, with 81 frames per video sub-sampled to 31 frames. We reconstruct with our pipeline to obtain $20$ different rigged meshes with $31$ poses each, for a total of $3.8$k rigged meshes and $115$k poses in the training set after filtering.
For the \texttt{cross-species} test set, we generate 32 unseen individuals, 15 from unseen species and 17 from species shared with the training set.
We include one rigged mesh per individual.

For the \texttt{single-species} training set, we select \texttt{Dog} as species and generate 20 images, corresponding to 20 individuals. Otherwise, the generation procedure matches the \texttt{cross-species} dataset, resulting in $400$ rigged meshes and $12.2$k poses after filtering. For the test set, we generate 2 unseen individuals
and one rigged mesh per individual.
Please see the supplement for additional details.

\begin{figure*}[t!]
    \centering
    \includegraphics[width=0.88\textwidth]{figures/feed_forward_gallery.pdf}
    \caption{
    \changed{\textbf{Zero-shot results on unseen individuals.} We show results of our feed-forward model on several unseen rigged meshes.}
    }
    \label{fig:feed_forward_gallery}
\end{figure*}

\begin{figure*}[t]
    \centering
    \includegraphics[width=\linewidth]{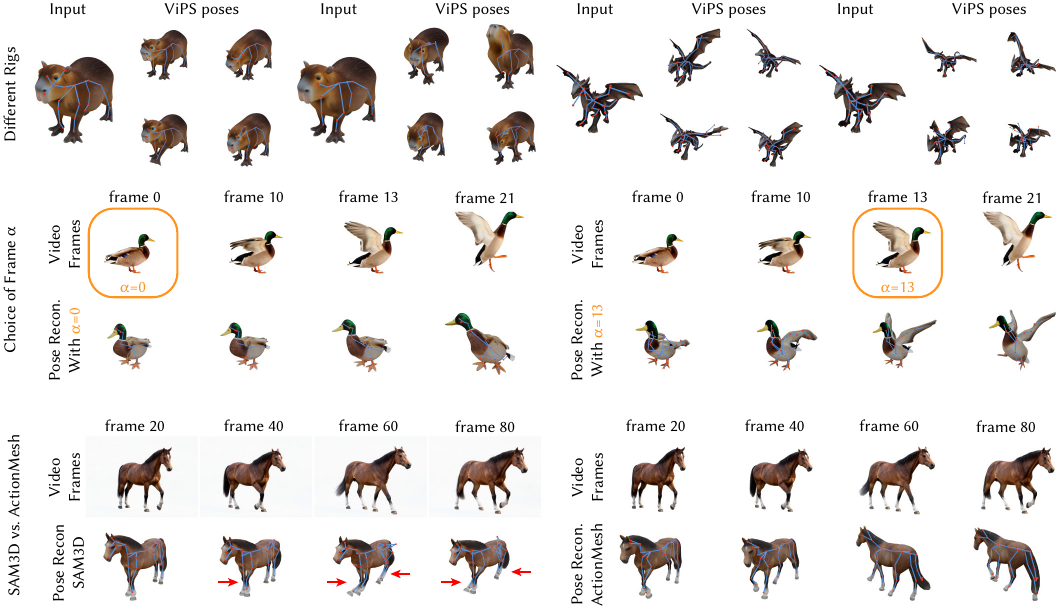}
    \caption{\textbf{Ablations of three choices.} 
    On top, we show samples from our feed-forward model using the same input mesh \emph{with different skeletons}, which has a limited effect on the generated poses. 
    In the middle, we show  the effect of using different frames $\alpha$ for the initial mesh $M_\alpha$, which can have a large effect on the quality of the reconstructed poses if there are significant topology changes of the 3D shape in different frames. At the bottom, we compare using Actionmesh~\cite{sabathier2026actionmeshanimated3dmesh} for 4D reconstruction to per-frame SAM3D~\cite{chen2025sam3d} compreconstruction. We can see that the unreliable 3D surface correspondences between per-frame SAM3D reconstructions significantly lower pose quality (reconstruction errors are marked with red arrows).}
    \label{fig:supp_ablation}
\end{figure*}

\subsection{Pose Sampling and Interpolation}
\label{sec:res_pose_sampling}

Our goal is to learn a pose distribution that is diverse and plausible, and can be applied to unseen skeletons, and even unseen species of animals without the need to re-train or fine-tune our model, and without the need for artist-created data. To evaluate how well our method achieved this goal, {we first show several samples from the pose space of our feed-forward model for unseen skeletons and then compare to several baselines, both qualitatively and quantitatively.}

{\paragraph{Pose samples for unseen rigged meshes with varying topology.} To evaluate if pose knowledge from the video generator was successfully distilled into our feed-forward model, Figure~\ref{fig:feed_forward_gallery} shows poses randomly sampled by our feed-forward model for $11$ rigged meshes in the \texttt{cross-species} test set. Even though exact skeleton topologies, mesh shapes and textures may be new to our feed-forward model, it generates poses that are plausible and natural for the given input object in just 0.5 seconds per pose.}

\paragraph{Baselines.} 
As the first \textit{feed-forward framework} for \textit{universal pose generation} trained exclusively on \textit{unlabeled generated video data}, \name lacks a direct end-to-end baseline. Consequently, we evaluate our components against state-of-the-art methods in related sub-tasks. We identify {AnyTop}~\cite{gat2025anytop} as the most pertinent baseline due to its shared objective of \texttt{cross-species} learning and its related motion space.

\begin{table}[t]
\caption{
\textbf{Quantitative Comparison on Pose Plausibility and Diversity} 
against state-of-the-art baselines on cross-species ($\texttt{test}_\text{cs}$) and single-species ($\texttt{test}_\text{ss}$) benchmarks. \name significantly outperforms AnyTop variants in \textit{Plausibility} and \textit{Diversity}, as measured by a user-study-based ranking (LSR) and distribution similarity (FSD) to the expensive offline method Puppeteer, despite those baselines having access to specialized artist-data. Notably, our \textit{universal feed-forward model} maintains superior performance even in zero-shot cross-species scenarios, matching the plausibility of optimization-based reconstruction (Puppeteer) while providing a controllable generative manifold. Here we report the runtime for 10 sampled poses. }
\label{tab:comparison}
\footnotesize %
\renewcommand{\arraystretch}{1.1}
\setlength{\tabcolsep}{5pt}
\begin{tabularx}{\linewidth}{r >{\centering\arraybackslash}X >{\centering\arraybackslash}X >{\centering\arraybackslash}X >{\centering\arraybackslash}X >{\centering\arraybackslash}X} %
     \toprule
     & \multicolumn{2}{c}{$\texttt{test}_\text{cs}$}
     & \multicolumn{2}{c}{$\texttt{test}_\text{ss}$} \\
     \cmidrule(l    r){2-3}
     \cmidrule(l){4-5}
     & LSR$\uparrow$
     & FSD$\downarrow$
     & LSR$\uparrow$
     & FSD$\downarrow$
     & Time \\
     \midrule
     AnyTop-motion         & -0.21          & 1.49 & -2.78         & 2.28 & 2.5 sec \\
     AnyTop-pose           & -1.65          & 1.04 & 0.01          & 1.02 & 4 sec \\
     \textbf{\name (ours)} & \textbf{-0.04} & \textbf{0.31} & \textbf{0.65} & \textbf{0.87} & 4 sec \\
     \midrule
     Puppeteer             & 1.90           & $\mathrm{N/A}$  & 2.12          & $\mathrm{N/A}$  & 30 min \\
     \bottomrule
\end{tabularx}
\vspace{-10pt}
\end{table}

\begin{table}[t]
\caption{
\changed{\textbf{Data User Study.} 
We compare the motion plausibility and reconstruction quality of data generated with our pipeline against state-of-the-art baseline Puppeteer~\cite{puppeteer} on 6 individuals of the \texttt{cross-species} test set, showing the average percentage of users that preferred our method or the baseline. \honglin{Note that the timing reported here is for video reconstruction in our data pipeline (feed-forward 4D reconstruction–based optimization) and Puppeteer (tracking/rendering-loss–based optimization), and is separate from our feed-forward model’s inference time. }}
}
\label{tab:data_user_study}
\footnotesize %
\renewcommand{\arraystretch}{1.1}
\setlength{\tabcolsep}{5pt}
\begin{tabularx}{\linewidth}{r >{\centering\arraybackslash}X >{\centering\arraybackslash}X >{\centering\arraybackslash}X}
    \toprule
    & Motion Plausibility$\uparrow$
    & Motion Reconstruction$\uparrow$
    & Time per Video$\downarrow$ \\
    \midrule
    Puppeteer                            & 19.2\% & 29.2\% & 30 min \\
    \textbf{\name Data Pipeline (ours)} & \textbf{80.8\%} & \textbf{70.8\%} & \textbf{9 min} \\
    \bottomrule
\end{tabularx}
\vspace{-10pt}
\end{table}

For a fair comparison, we evaluate two variants: (i) \textit{AnyTop-motion}, where we extract individual poses by randomly sampling 200 frames from 20 generated motion sequences; and (ii)~\textit{AnyTop-pose}, an architectural variant we modify to learn a {static distribution of poses} rather than sequences. Notably, unlike ours, AnyTop variants are trained on {artist-authored datasets} and require \textit{privileged information}; specifically, semantic node labels and per-joint motion statistics—not typically available for unseen species. Artist data are much higher in quality and semantic alignment; but hard to scale. 
We additionally provide these baselines with a \textit{``best-effort'' approximation} for out-of-distribution assets: using joint indices as labels, the rest pose as the mean, and VLM-derived statistics from the closest available species. Despite this additional structural and semantic  guidance, our results demonstrate superior adaptability. 

We also compare to {Puppeteer}~\cite{puppeteer} as an upper bound and a reference for pose quality. Puppeteer is an optimization-based method, that requires 30 minutes on average to generate a pose sequence. We randomly sample poses from 200 frames of 20 generated sequences per individual.
The supplemental provides further comparisons of our data pipeline to {Puppeteer}.

\paragraph{Metrics.}
We measure two main properties of generated poses: \textit{diversity} and \textit{plausibility}. Note that we have no clear ground truth distribution to compare to, since our data pipeline is part of the contribution we aim to evaluate. Instead, we use the much more expensive Puppeteer as neutral ground truth.
The {Fréchet Skeleton Distance} (FSD)~\cite{maiorca2022frechetmotiondistance,maiorca2023validatingfmd}, defined as the Fréchet distance between generated samples and Puppeteer samples, using concatenated normalized joint positions as feature vectors,
measures both \textit{diversity} and \textit{plausibility} -- a lower value means better alignment with the Puppeteer distribution.
For \textit{plausibility}, we additionally perform a user study where users are tasked with comparing the plausibility of pose sets generated by our method to baselines in two-alternative forced choice comparisons. We had 24 participants answering a total of 360 pairwise comparisons. To get a per-method score, we show the Luce Spectral Ranking~\cite{maystre2015fast} (LSR) computed from these comparisons.
See
the supplement for more details on both metrics.

\paragraph{Discussion.}
{We compare random samples from our pose space to baselines on 6 individuals ($\texttt{test}_\text{cs}$) from the \texttt{cross-species} test set (2 from unseen species, and 4 from species shared with the training set), and on 2 individuals ($\texttt{test}_\text{ss}$) from the \texttt{single-species} test set. For each individual, our method and each baseline generate 200 poses. Table~\ref{tab:comparison} provides a quantitative comparison and Figure~\ref{fig:qual_comparison} show a qualitative comparison on a few examples.} A much larger set of samples is provided in the supplementary webpage. Our samples remain plausible and display more pose diversity than the AnyTop baselines. We reach similar plausibility and diversity as the optimization-based Puppeteer, while only requiring a fraction of Puppeteer's generation time.
On unseen shapes, baselines either collapse to conservative joint configurations or exhibit artifacts (implausible poses) on unseen skeletons (and topology). Supplemental compares between our generated training data versus Puppeteer.

\subsection{Pose Editing}

We use the guidance mechanism (\Cref{sec:method}) as a form of semantically meaningful, automatic inverse kinematics. Users specify sparse joint constraints via drags (e.g., by moving a paw forward or raising the head), and guidance projects these edits back onto the learned manifold. The model therefore corrects the edit into a plausible pose while maintaining asset‑specific validity.

\paragraph{Constrained Edits.}
\Cref{fig:pose_editing} shows single‑handle edits. Each example includes the original pose, the edited pose, and the user-specified edit. In practice, the guidance behaves like automatic IK: it satisfies the handle motion but also redistributes motion across the chain in a semantically plausible way (e.g., shoulder and torso adjust when a paw is moved). Further, it imposes additional semantic constraints such as limiting extreme edits to a meaningful extent (e.g., head cannot be dragged too far below the feet). We further demonstrate smooth transitions between edits using our pose-space traversal between the edited poses. This produces coherent transitions between edits that are more plausible than joint‑space linear interpolation (see supplementary material).

\subsection{Controllable Video Generation}

Our pose space provides a simple interface for generating keyframes that can steer a video diffusion model. We select keyframes along a pose-space traversal (or between edits), render the corresponding mesh+skeleton proxy, and supply these as conditioning frames, see \Cref{fig:teaser}. 
This enables controllable, semantically aligned video generation: the video model is free to synthesize appearance and texture, while the pose sequence provides precise 3D control. See supplementary material for more examples. 

\changed{\subsection{Data Pipeline Evaluation}}

We evaluate our data pipeline independently of our feed-forward model by comparing videos reconstructed with our pipeline to videos reconstructed by Puppeteer~\cite{puppeteer} in a user study. 
20 participants were asked to compare motion plausibility and motion reconstruction accuracy
in a two-alternative forced choice test on 6 individuals from the \texttt{cross-species} test set, with 4 randomly selected motions shown for each individual. 
See Table~\ref{tab:data_user_study} and qualitative examples in Figure~\ref{fig:supp_data_comparison}. Both show that our pipeline tends to reconstruct motions more accurately, especially when long limbs are involved, which also results in improved motion plausibility.
See the supplement for user study details (App.~\ref{sec:suppl_user_study}).

We further analyze our data generation pipeline in Figure~\ref{fig:supp_artist_data_comparison}, Figure~\ref{fig:supp_different_video_priors} and Figure~\ref{fig:supp_ablation}. In Figure~\ref{fig:supp_artist_data_comparison}, we compare our reconstructed motions with artist-authored animations from open-source~\cite{objaverseXL} and commercial~\cite{trueboneszoo} datasets, showing that our generated motion data achieves quality comparable to artist-authored animations. In Figure~\ref{fig:supp_different_video_priors}, we show that our pipeline is not tied to a specific video prior: replacing Wan2.2-TurboDiffusion~\cite{TurboDiffusion2025} with other video priors yields similar plausible motions for the same character and prompt, suggesting that our method can benefit from future advances of video generative models. Please see App.~\ref{sec:suppl_data_comparison} for details.
In Figure~\ref{fig:supp_ablation} (middle and bottom row), we further conduct an ablation study on the effect of using different frames for the initial mesh reconstruction and the difference between our choice of feed-forward 4D construction and per-frame SAM3D reconstruction. Please see App.~\ref{sec:suppl_ablation} for details.

\section{Conclusion}
\label{sec:conclusion}

We have presented \name, a feed-forward framework that breathes life into raw character rigs by distilling video diffusion priors into a universal, semantically-aware pose manifold. By shifting the paradigm from scarce, artist-authored 4D motions to generative distillation, we bypass specialized datasets and move toward a truly species-agnostic foundation for animation. Whether applied to humans, quadrupeds, or mythical creatures, our approach infers plausible articulation directly from 2D data, enabling a single model to encode the diverse biomechanics of any asset. This bridge between unstructured generative priors and structured kinematic control redefines the 3D character as a steerable, differentiable proxy for the next generation of video synthesis.

\begin{wrapfigure}[5]{r}{0.15\linewidth}
	\raggedleft
    \vspace{-9pt}
	\hspace*{-0.7\columnsep}
	\includegraphics[width=\linewidth, trim={6mm 2mm 0mm 4mm}]{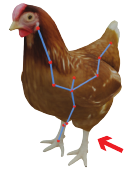}
	\label{fig:limitation_bad_skel}
\end{wrapfigure} 
\paragraph{Structural and Parameterization Constraints.} While \name effectively discovers the latent manifold of an existing rig, it remains \textit{inherently rig-dependent}, inheriting structural flaws like suboptimal bone placement or skinning weights from the input autorig, as illustrated by the chicken example in the inset. A  future direction is the joint feed-forward extraction of rig topology and pose space, similar to RigMo \cite{zhang2026rigmo}, or employing redundant representations where the network simultaneously predicts the deformed mesh, Gaussian splats, and skeletal parameters for cross-modal consistency. Performance on non-biological entities or complex insects remains sensitive to the underlying auto-rigs.

\paragraph{Prior Biases and Initial States.} Our reliance on video diffusion models and 4D extraction introduces an implicit data-bias; for highly specialized or \textit{rare motions} (e.g., specific flight patterns of \texttt{swifts}), current generative priors may still lack sufficient fidelity. Nevertheless, as shown in Figure~4 of the supplement, our data pipeline can be seamlessly integrated with different video models as priors, suggesting that it may continue to benefit from advances in video generative models. We also acknowledge the \textit{start pose} problem $M_\alpha$, where the quality of the learned space is sensitive to the alignment of the initial rest pose relative to the video sequence. Future work could address this by leveraging text-driven motion priors to anchor the pose space, allowing users to drive `walks' or specific gaits through natural language prompts rather than just latent interpolation. See supplemental for comparison on other video priors. 

\paragraph{Scaling Beyond Kinematics.} An immediate evolution is the transition from a pose space to a dynamic \textit{motion space}, explicitly modeling rig-parameter trajectories to capture gait families and long-horizon style modulation. Furthermore, we aim to extend the framework to handle complex deformation regimes, including elastic soft bodies (rubber/cloth), topology-changing phenomena (fracture/fluids), and secondary dynamics (jiggle/inertia). %

\section{Acknowledgements}
\label{sec:acknowledgements}

This work was supported in part by a Roblox Graduate Fellowship to Honglin Chen. We thank Inbar Gat and Sigal Raab for helpful discussions on AnyTop, Remy Sabathier for his help on ActionMesh code, and Zifan Shi for insightful discussions on RigAnything. Their support was invaluable to the development and success of ViPS.

\bibliographystyle{ACM-Reference-Format}
\bibliography{main}

\appendix

\clearpage
\appendix

\section{Overview}
\label{sec:suppl_overview}
In this supplementary material, we evaluate our feed-forward model separately from the data pipeline in Section~\ref{sec:suppl_model_comparison}, and the data pipeline separately from the feed-forward model in Section~\ref{sec:suppl_data_comparison}. Additionally, we evaluate different choices in our method with an ablation in Section~\ref{sec:suppl_ablation}. In Sections~\ref{sec:suppl_dataset} and~\ref{sec:suppl_data_pipeline} we provide additional details about the dataset we generated with out pipeline, and about the data pipeline itself, respectively, and Section~\ref{sec:suppl_user_study} provides additional detail about the user study.

The included supplementary website provides additional examples of applying \name to create keyframes for video generation, and shows several examples of walks through the pose space. Additionally, examples of videos and images from the data generation pipeline are provided, as well as additional qualitative comparisons.

\section{Model Comparison}
\label{sec:suppl_model_comparison}
\begin{figure*}[t]
    \centering
    \includegraphics[width=\linewidth]{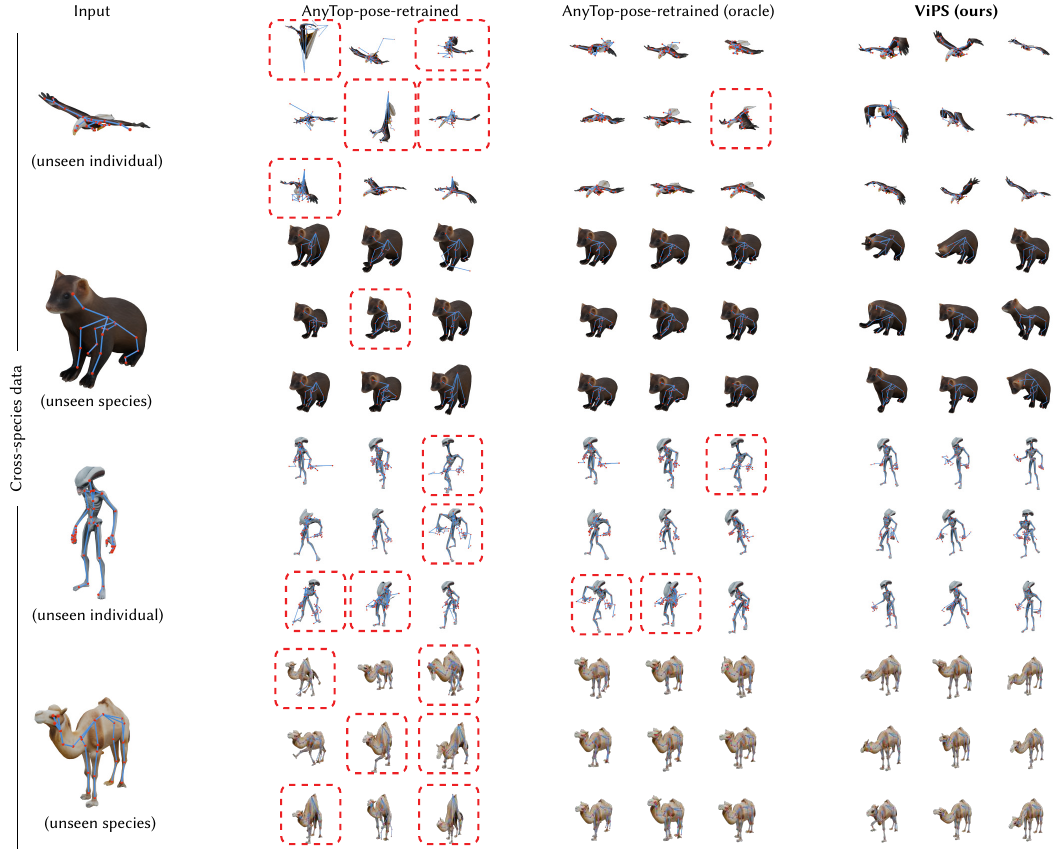}
    \caption{\textbf{Model comparison.} We evaluate our feed-forward model independently of the data pipeline by comparing poses generated by our method to poses generated by two variants of AnyTop-pose that both have been re-trained on the same data as our model. The \emph{oracle} variant receives ground truth node statistics and node labels as input that would not be available at inference time in practice and the other variant receives a ``best-effort'' approximation from data that could realistically be available at inference time. Even though our model does not require node statistics or node labels as input, it performs slightly better than the oracle variant, and significantly outperforms the best-effort variant.}
    \label{fig:supp_model_comparison}
\end{figure*}

To evaluate our feed-forward model independently of the data, we compare it to a version of AnyTop-pose that we re-trained on our data and call \emph{AnyTop-pose-retrained}.
Unlike AnyTop, our feed-forward model does not require node statistics or node labels as input, due to our use of per-node semantic features that are automatically computed from the rigged input mesh (see Section 3.2 of the main paper). During the training of \emph{AnyTop-pose-retrained}, as meaningful node labels (e.g., ``head'', ``neck'', ``hips'') are not available in our data, we use all zeros as node embeddings and the ground truth node statistics from the training data. As node statistics and node labels are not available at inference time in practice, we use the same ``best-effort'' approximation as in Table~\ref{tab:supp_model_comparison} for AnyTop-pose-retrained, using all zeros as node embeddings, the rest pose as mean, and VLM-derived statistics from the closest available species. We also include an additional variant of the baseline that we provide with ground truth node statistics as input, and that we call \emph{AnyTop-pose-retrained (oracle)} to emphasize that this ground truth input data is not available for inference in practice. Similar to our approach, we directly use the output of AnyTop, without additional regularization through inverse kinematics. Unlike in the Table~\ref{tab:supp_model_comparison} comparison, we do have ground truth in this comparison, therefore we compute the Frechet Skeleton Distance (FSD) relative to the ground truth. Additionally, we compute an overfitting measure based on the nearest-neighbor (NN) distances to the dataset samples that we denote as $O_\text{NN}$. Specifically, we use $200$ generated poses from each method and from the ground truth, and compute the ratio of the average ground truth intra-set NN distances (from each ground truth pose to the closest other ground truth pose) to the average inter-set NN distances (from each generated pose to the closest ground truth pose). Intuitively, inter-set distances that are smaller than ground truth intra-set distances indicate overfitting, as generated samples tend to be closer to the ground truth samples than other ground truth samples. Values $> 1$ indicate overfitting, while values $\leq 1$  indicate no overfitting, with very small values indicating misalignment between the distributions. A quantitative comparison is shown in Table~\ref{tab:supp_model_comparison} and a qualitative comparison in Figure~\ref{fig:supp_model_comparison}. We see that our model improves upon the best-effort version of AnyTop-pose-retrained and even slightly outperforms the oracle version, even though it does not require ground truth node labels or node statistics as input.

\begin{figure*}[t]
    \centering
    \includegraphics[width=\linewidth, height=2in]{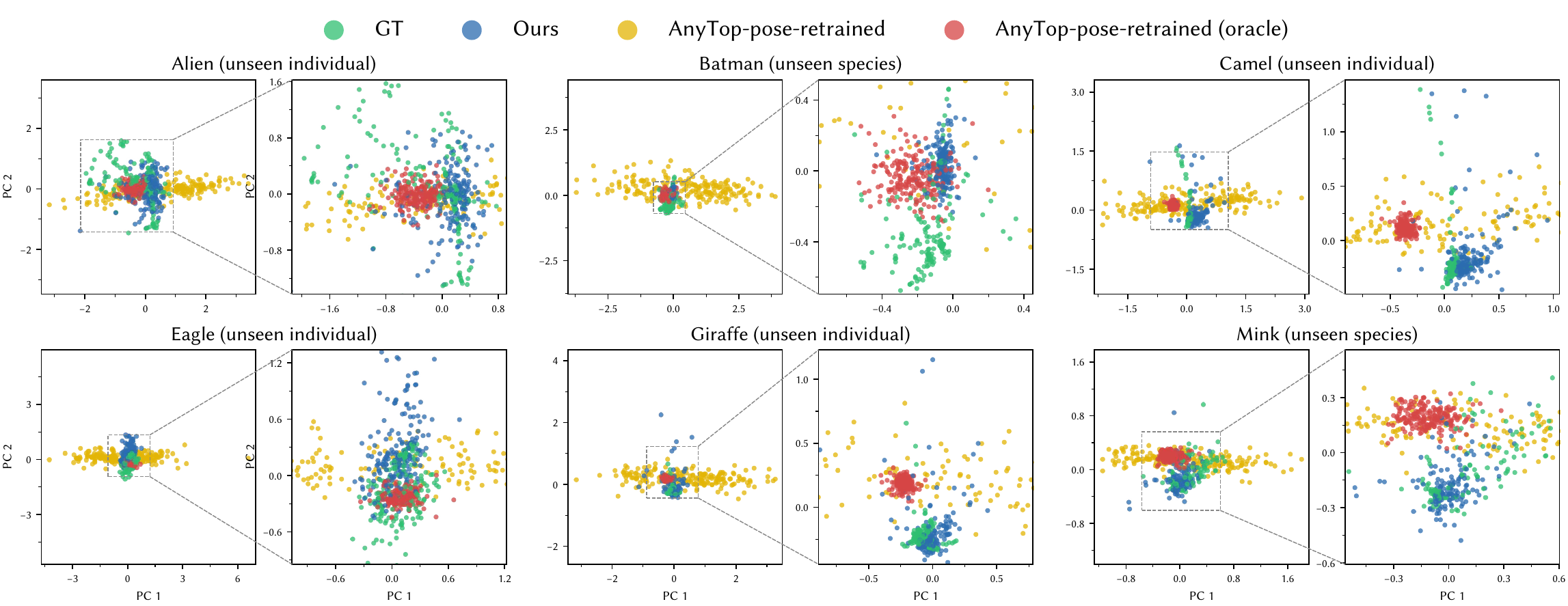}
    \caption{\textbf{Sample distributions.} We visualize pose distributions using the two first principal components of the concatenated skeleton node positions. We compare our poses (blue) and the two baselines (yellow and red) to the ground truth distribution (green) for several different individuals. Our distribution follows the ground truth distribution more closely than both of the baselines.}
    \label{fig:supp_model_distributions}
\end{figure*}

We additionally visualize a 2D embedding of pose distributions of our method compared to baselines and the ground truth for several individuals in Figure~\ref{fig:supp_model_distributions}. We use the concatenated node positions as feature vectors and show the first two principal components of the distributions as x- and y axes. The plots confirm that our generated pose distributions better match the ground truth than the baselines.

\begin{table}[t]
\caption{
\textbf{Comparison of Feed-Forward Models on Pose Plausibility and Diversity.} We compare our feed-forward model to two version of AnyTop trained on the same data as our model. One version uses the same best-effort approximation of the input we use in Table~\ref{tab:supp_model_comparison}, the other (oracle) uses inputs derived from the ground truth, and is thus not usable in practice, but uses the same type of input as in the original AnyTop paper~\cite{gat2025anytop}. We measure distribution similarity (FSD) to the ground truth distribution, and overfitting using a nearest-neighbor-based metric ($O_\text{NN}$). \name performs significantly better then the best-effort version of AnyTop-pose and even slightly outperforms the oracle version, without requiring ground truth data as input.}
\label{tab:supp_model_comparison}
\footnotesize %
\renewcommand{\arraystretch}{1.1}
\setlength{\tabcolsep}{5pt}
\begin{tabularx}{\linewidth}{r >{\centering\arraybackslash}X >{\centering\arraybackslash}X} %
     \toprule
     & FSD$\downarrow$ & $O_\text{NN}${\scriptsize(no overfitting if $\leq 1$)} \\
     \midrule
     AnyTop-pose           & 2.00 & \textbf{0.09}  \\
     AnyTop-pose (oracle)  & 0.43 & \textbf{0.22}  \\
     \textbf{\name (ours)} & \textbf{0.31} & \textbf{0.29}  \\
     \bottomrule
\end{tabularx}
\end{table}

\begin{figure*}[t]
    \centering
    \begin{minipage}[t]{0.39\linewidth}
        \centering
        \includegraphics[width=\linewidth]{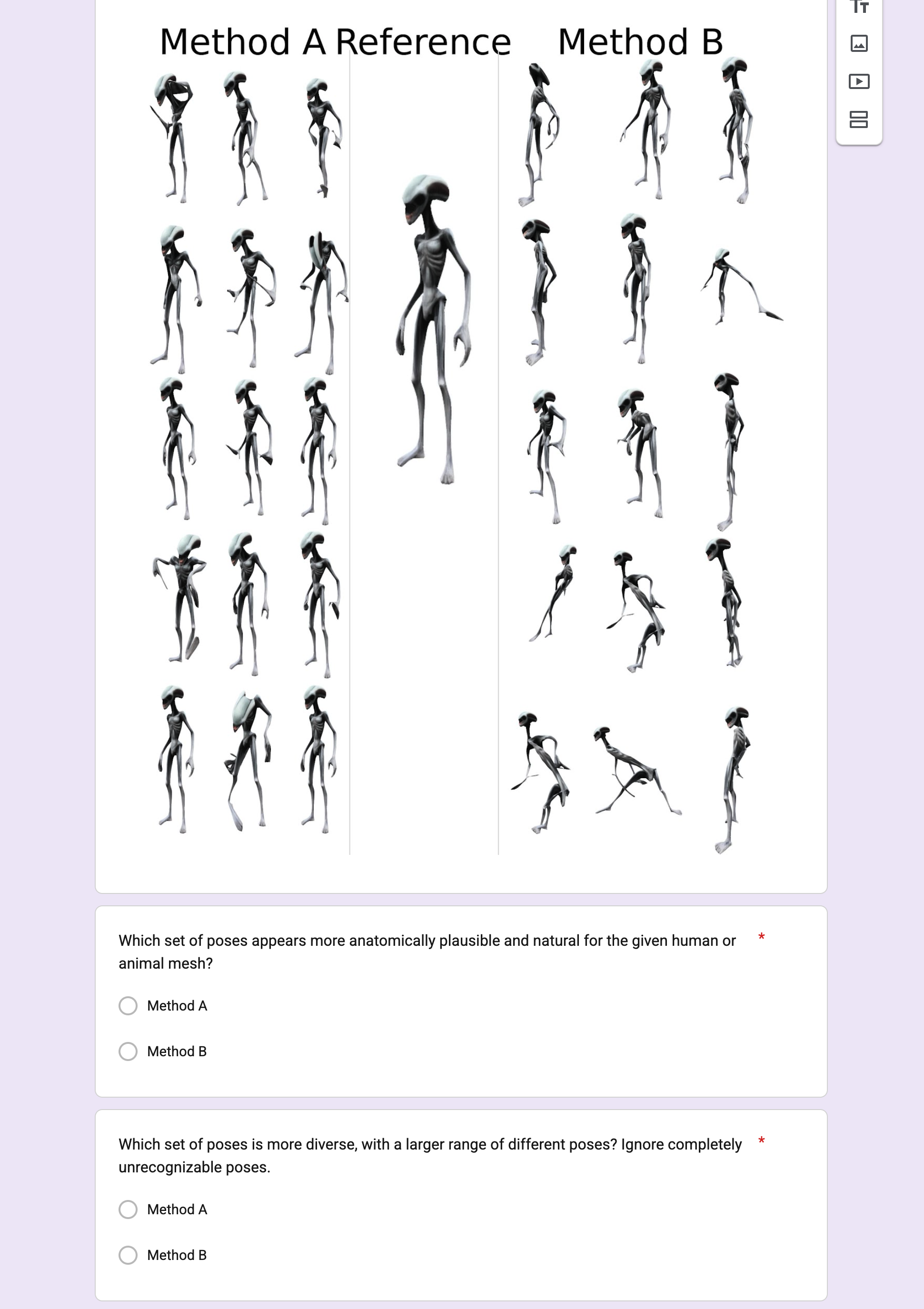}
    \end{minipage}
    \hfill
    \begin{minipage}[t]{0.59\linewidth}
        \centering
        \includegraphics[width=\linewidth]{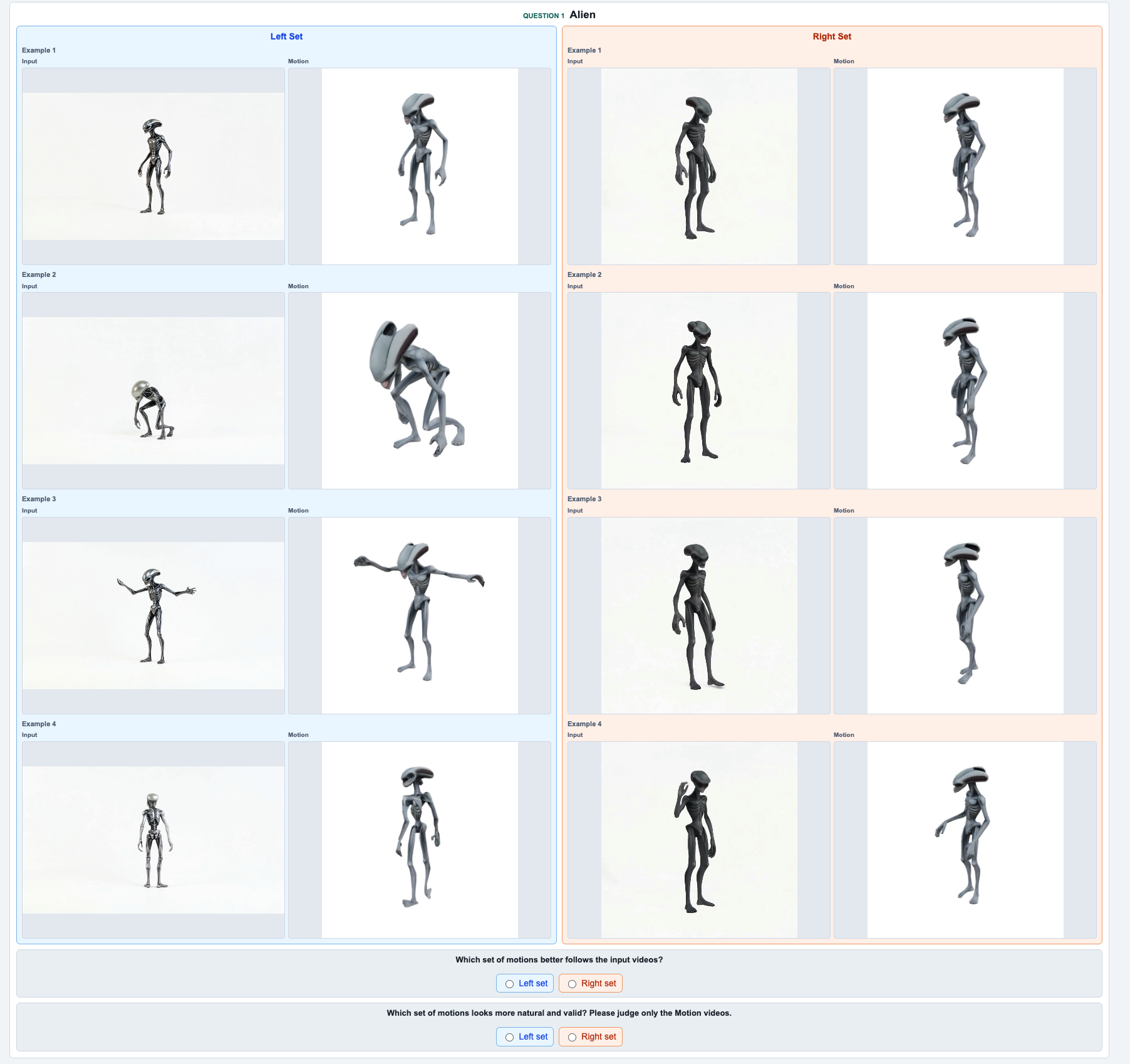}
    \end{minipage}
    \caption{\textbf{User study screenshots.} We compare pose distributions for a given individual with two-alternative forced choice tests. Left: the main user study interface, where we show $15$ poses of each method in grids on the left and right side, with the input individual in its rest pose in the center. \honglin{Right: the data pipeline user study interface, where we display $4$ set of input video and the corresponding reconstructed pose video for each of the two methods on the left and right-hand side.}}
    \label{fig:supp_user_study}
\end{figure*}

\section{Data Comparison}
\label{sec:suppl_data_comparison}
We compare our data pipeline to Puppeteer independently of the feed-forward model, we compare poses of our generated data and data generated by Puppeteer. Unlike Puppeteer, we do not attempt to explicitly track object points across the video when optimizing our poses to match the video frames, as this has proven to be brittle in earlier experiments. Instead, we rely on the 4D reconstructions from ActionMesh~\cite{sabathier2026actionmeshanimated3dmesh}, which have proven to be significantly more robust. Several examples are compared in Figure~\ref{fig:supp_data_comparison}. Our 4D-reconstruction-based approach tends to reconstruct poses more accurately than Puppeteer's tracking-based approach.

\honglin{
We further analyze the flexibility and motivation of our data generation pipeline in Figures~\ref{fig:supp_different_video_priors} and~\ref{fig:supp_artist_data_comparison}. Figure~\ref{fig:supp_different_video_priors} shows that our pipeline is not tied to a specific video model: by replacing our default Wan2.2-TurboDiffusion~\cite{TurboDiffusion2025} prior with alternative video generation models such as Kling~\cite{klingteam2025klingomnitechnicalreport}, Runway~\cite{runway2026}, and SeedDance~\cite{seedance2026seedance20advancingvideo}, we can obtain different plausible motion priors for the same input character and prompt. This suggests that our pipeline can naturally benefit from future improvements in video generation models. In Figure~\ref{fig:supp_artist_data_comparison}, we compare our reconstructed motions with artist-authored animations from both open-source~\cite{objaverseXL} and commercial sources~\cite{trueboneszoo}. While artist-authored data can provide high-quality motions, its quality and coverage vary substantially depending on the dataset source, asset origin, and animator effort. In contrast, our generated data pipeline provides a scalable way to produce diverse object-specific motions, reducing the dependence on expensive or sparsely available manually authored animation clips.
}

\section{Ablation}
\label{sec:suppl_ablation}
In this ablation, we examine the effect of three choices in our data pipeline and feed-forward model.
First, we show the effect of using different rigs generated by RigAnything for the same individual. Figure~\ref{fig:supp_ablation} (top) shows several examples. We can see that the exact rig topology does not strongly affect the rig plausibility or diversity, as rigs typically have sufficient degrees of freedom to accurately capture the pose distribution of a given individual, and our feed-forward model has been trained with a diverse enough set of rigs to generalize to diverse rig topologies.
Second, we show the effect of choosing different frames $\alpha$ for the rigged mesh $M_\alpha$ in Figure~\ref{fig:supp_ablation} (middle). This choice is important if the 3D shape reconstructed for different frames of the video have different topologies, such as the example we can see here of a duck with open and closed wings. Choosing a frame $\alpha$ with closed wings results in a mesh $M_\alpha$ that does not support opening the wings.
Finally, we compare our data pipeline to a version where we apply SAM3D~\cite{chen2025sam3d} in each frame to get per-frame mesh reconstructions instead of using Actionmesh~\cite{sabathier2026actionmeshanimated3dmesh}, and use the chamfer distance between surface samples in adjacent frames to establish correspondence. Figure~\ref{fig:supp_ablation} (bottom) shows examples where we can clearly see that a lack of reliable correspondence between 3D surface points makes pose optimization more difficult and less reliable, resulting in lower-quality poses. Specifically the legs of the horse are hard to track for a 2D tracker, due to their thin structure and frequent occlusions.

\section{Data Details}
\label{sec:suppl_dataset}
The full set of species we used in our \texttt{cross-species} training set is listed here:
\begin{itemize}
    \item \texttt{Bipeds}: 11 species (Cassowary, Chicken, Chimpanzee, Crane, Flamingo, Gorilla, Kangaroo, Ostrich, Penguin, Raptor, Trex)
    \item \texttt{Fantasy}: 10 species (Dragon, Fairy, Goblin, Griffin, Mermaid, Pegasus, Phoenix, Unicorn, Werewolf, Zombie)
    \item \texttt{Fish}: 7 species (Dolphin, Eel, Generic Fish, Sea Lion, Shark, Stingray, Whale)
    \item \texttt{Flying}: 9 species (Bat, Bird, Eagle, Goose, Owl, Parrot, Pigeon, Pteranodon, Seagull)
    \item \texttt{Humans}: 12 ``species'' (Alien, Angel, Astronaut, Baby, Child, Demon, Elder, Giant, Generic Human, Robot, Superhero, Wizard)
    \item \texttt{Quadrupeds}: 52 species (Anteater, Armadillo, Badger, Beaver, Bison, Camel, Capybara, Cat, Cheetah, Chinchilla, Cow, Crocodile, Deer, Dog, Donkey, Elephant, Fox, Frog, Gazelle, Giraffe, Goat, Hedgehog, Hippopotamus, Horse, Hyena, Koala, Komodo dragon, Leopard, Lion, Lizard, Llama, Lynx, Meerkat, Monkey, Moose, Mouse, Okapi, Otter, Pangolin, Pig, Porcupine, Raccoon, Rhinoceros, Sheep, Skunk, Squirrel, Tiger, Turtle, Wild boar, Wolf, Wombat, Zebra)
\end{itemize}
for a total of 101 species. As described in Section 4.1 of the main paper, for each species, we generate two images with Flux, corresponding to two different individuals of the species, and 20 videos for each individual, corresponding to 20 different motions that we sample poses from.
Examples of prompts, generated images, generated videos, and reconstructed poses from our data pipeline are provided in the included webpage.
The \texttt{cross-species} test set includes one new individual each from two unseen species (Mink, Batman) and one new individual each from four seen species (Alien, Camel, Eagle, Giraffe).

\section{Addtional Data Pipeline Details}
\label{sec:suppl_data_pipeline}

\subsection{Image Prompt Generation with InternVL}
\label{sec:suppl_image_instructions}
We instruct InternVL~\cite{wang2025internvl3_5} to create a diverse set of individuals for a given species with the following instruction:
\begin{lstlisting}[frame=tb, backgroundcolor = \color{Apricot}] 
You are a prompt engineer for a text-to-image model.

Goal: Write ONE high-quality prompt per image for the given object/species name (about 30-50 words), with a strong emphasis on FULL-BODY / FULL-OBJECT visibility.

Hard requirements (MUST follow):
1) EXACT SUBJECT: Use the provided object/species name as the subject; do NOT change it to a different species/object.
2) FULL-BODY / FULL-OBJECT (CRITICAL): The ENTIRE subject must be visible from end to end (e.g., head-to-toe / nose-to-tail / top-to-bottom). Absolutely NO cropping, NO cut-off limbs/ears/tail, NO partial framing. Keep the subject centered with generous margins.
3) WIDE SHOT (CRITICAL): Use a wide shot with the camera pulled back enough to guarantee the whole subject fits comfortably in frame, with extra space around it.
4) BACKGROUND: Solid pure white studio background. NO ground plane, NO visible floor line, NO horizon.
5) LIGHTING: Even, soft, shadowless studio lighting. Avoid harsh cast shadows; keep reflections minimal.
6) NO NEW ELEMENTS: Do NOT introduce any other objects, humans, animals, props, text, logos, watermarks, scenery, or clutter.
7) Use side or front three-quarter viewpoint; AVOID direct front or rear views.

Diversity requirements (across prompts for the same subject):
- Vary viewpoint (side/front three-quarter), slight tilt angle (especially for animals), and pose/orientation while ALWAYS keeping full-body / full-object visibility. No front or rear views.
- Vary plausible appearance details while staying true to the subject:
  - Animals: coat color/pattern, age (adult/juvenile), body build, ear/tail position, head direction.
  - Rigid objects: material (metal/plastic/wood/glass), color, finish (matte/gloss), wear level (new/used).
  - Articulated objects: configuration (folded/unfolded), steering angle, arm joint angles, etc.
- Keep the style consistent: photorealistic / realistic 3D render with physically based materials and high detail.

Output format:
- Output ONLY the final prompt line.
- No quotes, no numbering, no bullet points, no extra commentary.

Subject name: <...>
Category: <...>
Object type:: <...>
Generate <N> diverse prompts for this subject.
Return each prompt on its own line.
\end{lstlisting}

\subsection{Motion Prompt Generation with InternVL}
\label{sec:suppl_video_instructions}
To generate motion prompts, i.e. prompts for generating a video that starts with a previously generated image, we use the following instruction for InternVL~\cite{wang2025internvl3_5}:
\begin{lstlisting}[frame=tb, backgroundcolor = \color{Apricot}]
You are a prompt engineer. You will be given:
- an object appearance description (EXACT; do not change it),
- a target motion (EXACT; must be used),
- a camera constraint (EXACT; must be used).

Write a single high-quality video generation prompt.

Hard requirements (MUST follow):
1. **Use the Provided Motion EXACTLY:** The prompt MUST describe the target motion faithfully. Do NOT replace it with a different motion. Do NOT add extra actions beyond what is stated.
2. **Keep Appearance CONSISTENT:** Use the provided appearance description as-is. Do NOT invent new colors, accessories, markings, clothing, species, or extra objects.
3. **NO NEW ELEMENTS (VERY STRICT):** Do NOT introduce any new objects, humans, animals, props, accessories, text, logos, extra scenery items, or additional entities not explicitly present in the provided appearance description. If something is not mentioned in the appearance, it must not appear in the prompt.
4. **Background & Lighting:** Keep the background simple and stable, consistent with the provided appearance. No new items in the scene.
5. **Camera Behavior:** STRICTLY STATIC FULL BODY SHOT. Tripod-locked. The entire object visible at all times. **Single shot, no transitions, no cuts.** NO camera movement. Avoid words like "pan", "zoom", "track", "dolly", "close-up", "follow", "cut", "scene change", "transition", "montage".
6. **Natural Physics:** Motion should have realistic weight and timing appropriate to the object.
7. **Style Default:** If no style is specified, default to "Photorealistic, 4k, high fidelity."
8. **Length:** 60-90 words.

Output ONLY the final prompt. No quotes. No bullet points. No extra commentary.

Object (appearance, keep exact): <appearance_text>
Target Motion (must use exact): <motion_text>
Camera Constraint (must use exact): Static Full Body Shot (Tripod View).
The entire object must be visible. NO CAMERA MOVEMENT.
NO NEW ELEMENTS: Do not introduce any new objects, humans, or animals not present in the provided appearance description.
\end{lstlisting}

\noindent To obtain the object appearance, to be used as \texttt{appearance\_text}, we ask InternVL to describe the object in the image:
\begin{lstlisting}[frame=tb, backgroundcolor = \color{Apricot}]
Describe the object's appearance and the background in 10-20 words.
Be concrete: species/type, main colors, textures/materials, distinctive features, and the background setting. Do NOT describe motion. 
Do NOT introduce ANY new objects/humans/animals that are not visible.
\end{lstlisting}

\noindent And to obtain a description of an object motion, to be used in the \texttt{motion\_text}, we use the following instruction:
\begin{lstlisting}[frame=tb, backgroundcolor = \color{Apricot}]
List <N> motion names for a <species_name> using Title Case labels like game animation clips (examples: Walk; Run; Attack; Get Up; Bark; Jump; Lower Head; Scratch; Howl; Die).
Each item must be a SHORT label (2-4 words max, no punctuation, and the whole list should be pose-diverse: each label should correspond to a clearly different full-body pose.
Emphasize species-specific, full-body motions with clear pose changes.
Avoid trivial micro-actions (chewing, blinking, swallowing, ear flicks, lip licking, open/close beak), tail-only loops (tail wagging/swishing/flicking), and static/near-static actions (standing still, idling with no clear pose change).
Include a mix of idles/loops and one-off actions (e.g., Idle variants, Walk/Run loops, Jump, Fall, Get Up, attacks/defense).
Separate them with semicolons. Do not include camera directions. Use only English.
Output ONLY the <N> motion names. No quotes. No bullet points. No extra commentary.
\end{lstlisting}

\subsection{Pose Filtering and Rig Filtering}
\label{sec:suppl_pose_filtering}
In rare cases, the 4D reconstruction with ActionMesh~\cite{sabathier2026actionmeshanimated3dmesh} in our data generation pipeline 
can produce nearly static 4D reconstructions when it fails to recover meaningful motion. To handle these cases, we introduce a filter that measures the root-mean-square (RMS) joint displacement of each frame relative to the first frame, normalized by the mesh bounding-box diagonal. We mark a clip as static and exclude it if at least 90\% of frames have displacement smaller than a threshold (we pick 0.0015). We filter out 0.71\% of the clips in the dataset.

Additionally, RigAnything~\cite{liu2025riganything} may occasionally produce rigs with artifacts, resulting in suboptimal degrees of freedom for capturing deformed poses during reconstruction or feed-forward prediction. To mitigate this issue, we filter generated rigs usinIg mesh-containment and topology constraints. Specifically, for each bone, we sample points along the parent-child segment and mark the bone as invalid if either the fraction of its length lying outside the mesh is at least 50\% or its maximum distance from the mesh surface exceeds a threshold (we choose 0.1). A rig is accepted only if it contains no invalid bones and its parent graph is acyclic. For each mesh, we evaluate up to 10 random seeds and keep the first rig that satisfies these criteria; if none do, we retain the best-scoring candidate among all attempts for robustness.

\section{Additional User Study Details}
\label{sec:suppl_user_study}
In the first user study, we compare poses distributions with two-alternative forced choice tests, comparing two sets of randomly sampled poses for the same individual. We display $15$ poses for each of the two methods in $3 \times 5$ grids on the left and right-hand side of the comparison, and the rigged input mesh is shown in the center. An example of a comparison is shown in Figure~\ref{fig:supp_user_study} (left). 
\changed{
In the second user study, we compare motion reconstruction quality from our data pipeline and Pupputeer with two-alternative forced choice tests, comparing two sets of randomly sampled motions for the same individual.
We display $4$ set of input video and the corresponding reconstructed pose video for each of the two methods in $4 \times 2$ grids on the left and right-hand side of the comparison.
An example of a comparison is shown in Figure~\ref{fig:supp_user_study} (right).
}
Particpants for both studies were university students and vision/graphics researchers.

\bibliographystyle{ACM-Reference-Format}
\bibliography{main}

\end{document}